# Robust Triple-Matrix-Recovery-Based Auto-Weighted Label Propagation for Classification

Huan Zhang, Zhao Zhang, *Senior Member*, *IEEE*, Mingbo Zhao, *Senior Member*, *IEEE*, Qiaolin Ye, *Member*, *IEEE*, Min Zhang and Meng Wang, *Senior Member*, *IEEE*

*Abstract* — The graph-based semi-supervised label propagation algorithm has delivered impressive classification results. However, the estimated soft labels typically contain mixed signs and noise, which cause inaccurate predictions due to the lack of suitable constraints. Moreover, available methods typically calculate the weights and estimate the labels in the original input space, which typically contains noise and corruption. Thus, the encoded similarities and manifold smoothness may be inaccurate for label estimation. In this paper, we present effective schemes for resolving these issues and propose a novel and robust semi-supervised classification algorithm, namely, the triple-matrix-recovery-based robust auto-weighted label propagation framework (ALP-TMR). Our ALP-TMR introduces a triple matrix recovery mechanism to remove noise or mixed signs from the estimated soft labels and improve the robustness to noise and outliers in the steps of assigning weights and predicting the labels simultaneously. Our method can jointly recover the underlying clean data, clean labels and clean weighting spaces by decomposing the original data, predicted soft labels or weights into a clean part plus an error part by fitting noise. In addition, ALP-TMR integrates the auto-weighting process by minimizing reconstruction errors over the recovered clean data and clean soft labels, which can encode the weights more accurately to improve both data representation and classification. By classifying samples in the recovered clean label and weight spaces, one can potentially improve the label prediction results. The results of extensive experiments demonstrated the satisfactory performance of our ALP-TMR.

## I. Introduction

In numerous emerging applications of pattern classification and data mining, real-world data typically contain noise and complex distributions and are not easy to distinguish due to lack of prior information [25][37-40][57-60], such as label information. However, real data typically have intrinsic special structures, and inter-class samples typically lie in different subspaces, which provides the possibility of distinguishing them using data classification. Since labeling numerous real data is difficult and costly and unlabeled samples are easier to obtain, semi-supervised classification (SSC), which aims at classifying samples using partially labeled data, has been attracting substantial attention, especially in the case that the number of labeled samples is limited [17-20][31][41-45][52-53]. As a classical graph-based SSC algorithm [61-70], label propagation (LP) [1-9] has been widely utilized due to its effectiveness and efficiency. LP models predict the labels of samples by propagating label information from labeled data to unlabeled data using their geometric structures and initial state [1-9], namely, by balancing the label fitness and the manifold smoothness.

Representative transductive LP algorithms mainly include Gaussian fields and harmonic function (GFHF) [1], linear neighborhood propagation (LNP) [2], special LP (SLP) [3], learning with local and global Consistency (LLGC) [5], projective label propagation (ProjLP) via label embedding [6], prior-class-dissimilarity-based LNP (CD-LNP) [4], sparse neighborhood propagation (SparseNP) [8], positive and negative label propagation (PN-LP) [7], and adaptive neighborhood propagation (AdaptiveNP) [9]. GFHF, LLGC, LNP and SLP optimize similar objective functions to predict the labels of samples by receiving information partly from the neighborhood and partly from the initial state. To encode the local neighborhood accurately, CD-LNP linearly reconstructs each data item by integrating class information to define the dissimilarity and discriminative neighborhoods [4]. PN-LP extends regular LP to negative LP [7]. AdaptiveNP improves the classification performance by integrating sparse coding and neighborhood propagation [30], namely, the reconstruction and classification errors are jointly minimized, which differs from most LP algorithms, which explicitly separate the construction of graph weights from the LP step.

Although enhanced results have been delivered by the LP methods that are discussed above, they still drawbacks that can degrade the classification results. First, the steps of label prediction and weight assignment are typically conducted in the original input space, namely, the labels and local neighborhood of each sample are encoded directly over the original data. However, real-world data typically have various noises, unfavorable features and even corruptions; hence, it is likely that the estimated labels and encoded similarities are inaccurate in practice. Second, for single-label classification (one label is assigned for each sample), it is preferable for the estimated soft label vector of each sample to contain only one nonzero entry, which determines its class assignment. However, the estimated labels typically have unfavorable mixed signs and noise of various levels, which degrades the results [6] [8]. Thus, the labels that are determined over the original predicted soft labels may be inaccurate or even completely incorrect; LLGC, LNP, GFHF, SLP, CD-LNP, PN-LP and AdaptiveNP suffer from this problem. To eliminate the negative effects of noise and mixed signs in the estimated soft labels, SparseNP [8] and ProjLP [6] were introduced as feasible approaches. ProjLP can predict more discriminating "deep" soft labels of data by using label embedding to reduce

---

H. Zhang and M. Zhang are with School of Computer Science and Technology, Soochow University, Suzhou, China (e-mails: huan241125@163.com, minzhang@suda.edu.cn)

Z. Zhang is with the Key Laboratory of Knowledge Engineering with Big Data (Ministry of Education), Hefei University of Technology, Hefei, China; was also with School of Computer Science and Technology, Soochow University, Suzhou 215006, China (e-mail: cszzhang@gmail.com)

M. Wang is with the Key Laboratory of Knowledge Engineering with Big Data (Ministry of Education), Hefei University of Technology, Hefei, China (e-mail: eric.mengwang@gmail.com)

Q. Ye is with School of Computer Science, Nanjing University of Science and Technology, Nanjing 210094, China (e-mail: yqlcom@njfu.edu.cn)

M. Zhao is with the Department of Electronic Engineering, City University of Hong Kong, Kowloon, Hong Kong (e-mail: mzhao4-c@my.cityu.edu.hk)

the noise and mixed signs in soft labels to enhance the classification performance [6]. In contrast to ProjLP, SparseNP explicitly employs a sparse regularization on the predicted soft labels for the joint optimization to reduce the unfavorable mixed signs and the noise in the soft labels on the classification result [8]. In this paper, we develop another method and propose a simple yet effective approach for removing noise and mixed signs from the predicted soft labels via clean label recovery. Third, SLP, LNP, CD-LNP, LLGC, GFHF, ProjLP, SparseNP and PN-LP typically suffer from a performance-degrading issue that is caused by constructing the graph weights prior to the LP step, namely, by pre-calculating the weights over the original data in a separable step but failing to ensure that the pre-obtained weights are jointly optimal for the subsequent label estimation, thereby resulting in inaccurate predicted labels [19][23][26][36]. In addition, the widely used Gaussian kernel weights in [1][3][5][28] and local reconstruction weights in [2][4][6][8] require the selection of an optimal number of nearest neighbors or kernel width, which is never easy in practice. AdaptiveNP resolves this issue by integrating the adaptive weight construction and neighborhood propagation into a unified model; however, it still operates on the original data and label spaces. Therefore, the presence of both noise and mixed signs may render its prediction result fragile even if the corruption level is low. The pre-calculated weights and jointly computed weights from previous work may still suffer from lack of ensured robustness for data representation since without proper constraints on the weight matrix, they may contain incorrect inter-class connections that directly result in erroneous label prediction results and inaccurate similarities. Thus, it will be better to reduce the incorrect inter-class connections in weights and to use them to encode the neighborhood and manifold smoothness for enhancing the label prediction.

Therefore, in this paper, we propose effective strategies for addressing the drawbacks that are discussed above, and we propose a novel LP model that is robust to the noise and mixed signs in the original data, estimated soft labels and weights. The main contributions are summarized as follows:

(1) A novel triple-matrix-recovery-driven robust auto-weighted label propagation model, which is referred to as ALP-TMR, is proposed for semi-supervised classification. ALP-TMR introduces a novel triple-matrix-recovery-based mechanism for enhancing the robustness of the learning process to the noise and unfavorable mixed signs in the data, predicted soft labels and jointly computed weights.

(2) To discover the underlying clean label and clean data spaces, ALP-TMR presents a simple and intuitive approach in which the original data/estimated soft label matrix is factorized into a recovered clean data/clean label matrix and a sparse error. Then, for accurate classification, the final labels of samples are calculated in the recovered label space.

(3) To define the locality and manifold smoothness, ALP-TMR encodes the similarity more accurately by recovering the underlying clean weights. ALP-TMR decomposes the original weight matrix into a recovered clean weight matrix with fewer inter-class connections and a sparse error that is fitted to the incorrect connections. To make the recovered weights jointly optimal for representing and classifying data, ALP-TMR integrates the adaptive weighting by jointly minimizing the reconstruction errors based on recovered clean samples, clean labels and clean weights. The auto-weighting can also avoid the tricky issue of selecting the neighborhood size or the kernel width when assigning weights.

The remainder of this paper is organized as follows: In Section II, we briefly review the related work. Section III presents the formulation, optimization and convergence analysis of our ALP-TMR method. Section IV describes the experimental results and analysis. Finally, the conclusions of this work are presented in Section VI.

## II. RELATED WORK

### A. Regular Transductive Label Propagation

Consider a collection of samples $X = [X_L, X_U] \in \mathbb{R}^{n \times N}$, where $X_L = [x_1, x_2, ..., x_l] \in \mathbb{R}^{n \times l}$ is a labeled set, in which each sample $x_i \in \mathbb{R}^n$ is associated with a unique label $i (1 \le i \le c)$, where $c$ is the number of classes and $n$ is the dimensionality of the original data. $X_U = [x_{l+1}, x_{l+2}, ..., x_{l+u}] \in \mathbb{R}^{n \times u}$ is an unlabeled set, namely, $l + u = N$ is the total number of samples. Then, transductive LP methods aim at propagating label information of labeled set $X_L$ to unlabeled set $X_U$ based on their intrinsic relations, which are encoded by a pre-constructed weighted neighborhood graph $G = (X, E)$. To assign weights $W_{i,j}$ for measuring the pairwise similarities between samples, Gaussian kernel weights [1][3][5] and the reconstruction weights [2][4][6][8] are widely used in available methods. Let $Y = [y_1, y_2, ..., y_{l+u}] \in \mathbb{R}^{c \times (l+u)}$ denote the initial label set of all samples, where $y_i \in \mathbb{R}^c$ is a column vector, in which $y_{i,j} = 1$ if $x_j$ is labeled as $i (1 \le i \le c)$ and $y_{i,j} = 0$ for unlabeled data $x_j$. Denote by $F = [f_1, f_2, ..., f_{l+u}] \in \mathbb{R}^{c \times (l+u)}$ the estimated soft label matrix. A general framework for most available transductive LP algorithms can be defined as

$$\min_F tr(FLF^T) + tr((F-Y)U(F-Y)^T), \quad (1)$$

where $L$ is graph Laplacian matrix $L = D - W$ or normalized graph Laplacian matrix $L = D^{-1/2}(D-W)D^{-1/2} = I - D^{-1/2}WD^{-1/2}$, $W = [W_{i,j}] \in \mathbb{R}^{(l+u) \times (l+u)}$ is the weight matrix, $D$ is a diagonal matrix with entries $D_{ii} = \sum_j W_{i,j}, \forall_i$, $tr(\bullet)$ is the trace operator, $^T$ is the transpose of the corresponding matrix, and $U$ denotes a diagonal matrix for balancing the similarity-based manifold smoothness and label fitness. The label of each sample $x_i$ is received partially from its neighbors and partially from the initial state [1-9]. Matrix $U$ includes the weighting factors for the labeled and unlabeled data, which is defined as

$$U = \begin{bmatrix} U^l & 0 \\ 0 & U^u \end{bmatrix} \in \mathbb{R}^{(l+u) \times (l+u)}, \quad (2)$$

where $U^l = diag(u_i^l)$ and $U^u = diag(u_i^u)$ are diagonal matrices with $u_i^l, i = 1, ..., l$, and $u_i^u, i = l+1, ..., l+u$, respectively, as the entries for weighting the labeled and unlabeled data.

### B. Robust Label Propagation: SparseNP and ProjLP

We briefly review the formulations of SparseNP and ProjLP that consider eliminating the negative effects of unfavorable mixed signs in the predicted soft labels for classification.

**SparseNP.** SparseNP attempts to reduce the mixed signs and noise in the predicted labels via sparse regularization. SparseNP improves the discrimination performance of the predicted labels $F$ by imposing the sparse $l_{2,1}$-norm on $F$ directly for joint optimization [8]. By assuming that the data space and label space share the same reconstruction relations, the objective function of SparseNP is defined as

$$\min_F \|F - FW^T\|_F^2 + tr((F-Y)UD(F-Y)^T) + \psi\|F^T\|_{2,1}, \quad (3)$$

where $\|F - FW^T\|_F^2$ is the label reconstruction error, $D$ is a diagonal matrix with entries $D_{ii} = \sum_j W_{i,j}$, $\|F^T\|_{2,1}$ is the $l_{2,1}$-norm-based soft label matrix and $\psi$ is a tuning parameter. Due to the use of the $l_{2,1}$-norm, rows of the estimated soft label matrix $F^T$ will shrink to zeros theoretically [24] [29][34], which can potentially encourage the soft labels to be discriminating to enhance the classification result.

**ProjLP**. ProjLP is another effective approach for reducing the mixed signs and noise in the estimated soft labels by computing the discriminative "deep" labels of the samples. ProjLP explicitly obtains an $l_{2,1}$-norm-based robust projection $P = [p_1, p_2, ..., p_d] \in \mathbb{R}^{(c+1) \times (c+1)}$ for converting the original "shallow" label vector $f_i$ into a deep label vector via the label embedding in terms of $P^T f_i$. The optimization problem of the recently proposed ProjLP algorithm is formulated as

$$\min_{F,P} \|P^T F - P^T F W^T\|_F^2 + tr\left((P^T F - Y) U (P^T F - Y)^T\right) + \alpha \|P\|_{2,1} \quad (4)$$

where $W$ is a weight matrix, $W_{i,j}$ are the entries of $W$, $P^T F$ is the "deep" soft label matrix that is obtained by embedding over $F$ and $\|P\|_{2,1}$ denotes the $l_{2,1}$-norm-based sparse projection. $\|F - FW^T\|_F^2$ denotes the reconstruction error that is based on $P^T F$ and $tr\left((P^T F - Y) U (P^T F - Y)^T\right)$ is the label fitness error between $P^T F$ and initial label $Y$. Due to the joint computation of $P$, the embedded labels $P^T F$ will be more discriminating than the original soft labels in $F$ because the $l_{2,1}$-norm on $P$ can ensure that the learnt $P$ is sparse in rows, and the sparse embedding procedure can also remove unfavorable mixed signs from $P^T F$ for more accurate label estimation.

## III. ROBUST AUTO-WEIGHTED LABEL PROPAGATION VIA TRIPLE MATRIX RECOVERY (ALP-TMR)

### A. Main Notations that are Used in the Paper

First, some important notations are introduced. For any matrix $S = [s_1, s_2, ..., s_q] \in \mathbb{R}^{m \times q}$, its $l_{r,p}$-norm is defined as

$$\|S\|_{r,p} = \left[\sum_{i=1}^m \left(\sum_{j=1}^q |S_{i,j}|^r\right)^{p/r}\right]^{1/p}. \quad (5)$$

If $p=r=2$, it becomes the Frobenius norm; if $p=1$ and $r=2$, it is the $l_{2,1}$-norm [29][35]. Let $s^i$ be the $i$-th row vector of $S$, and let $S_i$ be the $i$-th column vector of $S$. Then, the Frobenius norm and the $l_{2,1}$-norm of $S$ are formulated as

$$\|S\|_F^2 = \sum_{i=1}^m \sum_{j=1}^q S_{i,j}^2 = \sum_{i=1}^m \|s^i\|_2^2 = tr(S^T S) = tr(SS^T)$$
$$\|S\|_{2,1} = \sum_{i=1}^m \left[\sum_{j=1}^q S_{i,j}^2\right]^{1/2} = \sum_{i=1}^m \|s^i\|_2 = 2tr(S^T O S) \quad (6)$$

where $O$ is a diagonal matrix with entries $O_{ii} = 1/2\|s^i\|_2$ and $S_{i,j}$ is the $(i,j)$-th entry of $S$. In practice, $\|s^i\|_2$ could be equal to zero. Under special cases, we can use the simple regularization approach to define $O_{ii} = 1/(2\|s^i\|_2 + \tau)$, where $\tau$ is a small number, so that $2\|s^i\|_2 + \tau$ can approximate $2\|s^i\|_2$. $I$ is an identity matrix and its dimensionality is automatically compatible. The horizontal (respectively, vertical) concatenation of a collection of matrices or vectors along a row (respectively, column) is represented using $[s^1; s^2; ...; s^m]$ (respectively, $[s_1, s_2, ..., s_q]$).

### B. The Objective Function

The formulation of our ALP-TMR is described. ALP-TMR focuses on enhancing the transductive label estimation and representation performances by improving the robustness to noise and outliers in the specified data $X$, estimated soft labels $F$ and constructed weights $W$, which operates by performing triple matrix decompositions on $X$, $F$, and $W$ simultaneously, namely, $X = X + E_X$, $F = F + E_F$ and $W = W + E_W$, where $X$, $F$ and $W$ are the recovered clean data, the clean soft labels and the clean weights, respectively, and $E_X$, $E_F$ and $E_W$ represent the data recovery error, the label recovery error and the neighborhood recovery error, respectively. The recovery errors $E_X$, $E_F$ and $E_W$ are all modeled by the robust $l_{2,1}$-norm for handling the gross sparse errors and outliers [19][29][30][32], which can make the computation of reconstructions $\|F^T - F^T\|_{2,1}$, $\|X^T - X^T\|_{2,1}$ and $\|W - W\|_{2,1}$ potentially robust to the noise and errors in $X$, $F$, and $W$. ALP-TMR also minimizes the reconstruction errors based on recovered clean labels $F$ and clean data $X$, which seamlessly integrates the robust LP and auto-weighting. These discussions lead to the following unified model for ALP-TMR:

$$\min_{F,W,E_F,E_X,E_W} J(F,X,W) + tr\left((F-Y)U(F-Y)^T\right)$$
$$+ \alpha\left(\|E_F^T\|_{2,1} + \beta\|E_X^T\|_{2,1} + \gamma\|E_W\|_{2,1}\right), \quad (7)$$
$$s.t.\ F = F + E_F,\ X = X + E_X,\ W = W + E_W$$
$$W, F \geq 0,\ W_{ii} = 0,\ e^T W = e^T$$

where $W, F \geq 0$ are nonnegative constraints, which enable the learnt weights to hold nonnegativity properties. The constraint $W_{ii} = 0$ avoids the trivial solution $W = I$ and $e^T W = e^T$ is the sum-to-one constraint, which is imposed to retain the geometrical properties of the learned weights. $J(F,X,W)$ is the joint reconstruction error over the recovered clean data $X$, clean labels $F$ and clean weights $W$, which is defined as

$$J(F,X,W) = \|F - FW\|_F^2 + \|X - XW\|_F^2 + \|W\|_F^2. \quad (8)$$

The weight matrix $W$ is jointly optimized adaptively. Although the above reconstruction error is similar to that of AdaptiveNP in form for weighting, the reconstruction weight matrix $W$ is simultaneously computed in the recovered clean data space $X$ and clean soft label space $F$, while the weight matrix in AdaptiveNP is still obtained based on the original data $X$ and soft labels $F$, which typically contain various noises and unfavorable mixed signs that cause inaccurate similarities. In addition, the jointly obtained weights $W$ ensure that the learned weights explicitly improve both the data representation and classification performances [18][20].

By substituting $F = F - E_F$, $X = X - E_X$ and $W = W - E_W$ into Eq. (7) and re-expressing the sum-to-one constraint $e^T W = e^T$, we obtain the following optimization problem:

$$\min_{F,W,E_F,E_X,E_W} \mathbb{Q}(F,W,E_F,E_X,E_W)$$
$$+ tr\left[\left((F-E_F)-Y\right)U\left((F-E_F)-Y\right)^T\right]$$
$$+ \alpha\left(\|E_F^T\|_{2,1} + \beta\|E_X^T\|_{2,1} + \gamma\|E_W\|_{2,1}\right), \quad (9)$$
$$s.t.\ (F - E_F) \geq 0,\ W \geq 0,\ W_{ii} = 0$$

where the joint reconstruction error term $J(F,X,W)$ can be reformulated as $\mathbb{Q}(F,W,E_F,E_X,E_W)$:

$$\mathbb{Q}(F,W,E_F,E_X,E_W) = \|H - H(W - E_W)\|_F^2 + \|(W - E_W)\|_F^2, \quad (10)$$

where $H = \left((F-E_F)^T, (X-E_X)^T, e^T\right)^T$. Matrix recovery over the samples and weights is alternately performed during the optimizations. Our ALP-TMR is more general and flexible for semi-supervised classification since it can be applied in both clean and noisy cases. For the ideal case that the data are absolutely clean, the predicted soft labels are correct and the encoded weights are accurate, and ALP-TMR can be reduced to the traditional setting.

The optimization procedures of our ALP-TMR method are performed alternately via the following three steps:

**(1) Robust auto-weighting in the recovered clean data space and clean label space:**

We show how to obtain the adaptive reconstruction weight matrix $W$ and error term $E_W$. Given the recovered clean soft labels $F = F - E_F$ and clean data $X = X - E_X$, we can obtain $W$ and $E_W$ from the following formulation:

$$\min_{W, E_W} \left\| \begin{pmatrix} F \\ X \\ e^T \end{pmatrix} - \begin{pmatrix} F \\ X \\ e^T \end{pmatrix} (W - E_W) \right\|_F^2 + \left\| (W - E_W) \right\|_F^2 + \alpha \gamma \left\| E_W \right\|_{2,1}. \quad (11)$$

$$s.t. \, W \geq 0, W_{ii} = 0$$

The adaptive reconstruction error encoded by $W$ is simultaneously computed based on recovered clean data and clean label spaces. This process can improve the representation and prediction results. Once the recovered adaptive weight matrix $W$ has been obtained, we can easily use it to recover the clean soft labels and the label fitness error.

**(2) Robust adaptive classification in the recovered clean label space via matrix decomposition:**

When the clean weight matrix $W$ has been recovered, we can use it to characterize the manifold smoothness, to preserve the locality between data points adaptively in the label space, and to propagate the label information of labeled data to unlabeled data. We can recover $F$ from

$$\min_{F, E_F} \left\| (F - E_F) - (F - E_F)(W - E_W) \right\|_F^2 + \alpha \left\| E_F^T \right\|_{2,1}$$
$$+ tr \left[ ((F - E_F) - Y) U ((F - E_F) - Y)^T \right] \quad . \quad (12)$$
$$s.t. \, (F - E_F) \geq 0$$

From the computed label matrix $F$ and the label fitness error $E_F$, the clean soft labels can be recovered as $F = F - E_F$.

**(3) Robust clean data recovery and error correction via matrix decomposition:**

We discuss how to compute the noise in $X$. Given the clean weights $W$, we can recover the clean data $X$ and $E_X$ via

$$\min_{E_X} \left\| (X - E_X) - (X - E_X)(W - E_W) \right\|_F^2 + \alpha \beta \left\| E_X^T \right\|_{2,1}. \quad (13)$$

After the sparse error $E_X$ for the above problem has been obtained, the clean data can be recovered as $X = X - E_X$. Then, we can use the clean data $X$, together with the recovered labels, to further update the clean adaptive weight matrix.

*C. Optimization*

We present the optimization procedure of ALP-TMR. Since the objective function of ALP-TMR in Eq. (9) has several variables and they depend on one another, it cannot be solved directly. Thus, we follow the common procedures to solve the problem of ALP-TMR alternately. First, we initialize $E_F$, $E_W$ and $E_X$ to zero matrices and initialize the weight matrix $W$ using the reconstruction weights of [10]. Then, the minimization of the problem is conducted as follows:

**1) *Given others, update the recovered clean data:***
First, we demonstrate how to recover the sparse error $E_X$ from $X$. The formula for solving $E_X$ is Eq. (13). Let $Q$ be a diagonal matrix with entries $q_{ii} = 1/(2 \| b^i \|_2)$, where $b^i$ is the $i$-th column vector of $E_X$. Suppose that each $b^i \neq 0$. Since $\| E_X^T \|_{2,1} = 2tr(E_X Q E_X^T)$, Eq.(13) can be approximated as

$$\min_{E_X} \left\| (X - E_X) - (X - E_X)(W - E_W) \right\|_F^2 + \alpha \beta \left\| E_X^T \right\|_{2,1}$$
$$= tr \left( (X - E_X) A (X - E_X)^T \right) + 2\alpha \beta tr \left( E_X Q E_X^T \right) \quad , \quad (14)$$

where $A = (I - (W - E_W))(I - (W - E_W))^T = (I - W)(I - W)^T$ is an auxiliary matrix. By differentiating Eq. (14) *with respect to $E_X$*, we can obtain $(E_X)_{t+1}$ at the $(t+1)$-th iteration as

$$(E_X)_{t+1} = X A_t (A_t + 2\alpha \beta Q_t)^{-1}, \, A_t = (I - W_t)(I - W_t)^T. \quad (15)$$

After $(E_X)_{t+1}$ has been obtained, we can update $Q_{t+1}$ as

$$Q_{t+1} = diag\left( (q_{ii})_{t+1} \right), where \, (q_{ii})_{t+1} = 1/\left[ 2 \| b_{t+1}^i \|_2 \right], \, \forall i, \quad (16)$$

where $b_{t+1}^i$ denotes the $i$-th column of $(E_X)_{t+1}$. We initialize the diagonal matrix $Q$ as an identity matrix. As we have $(E_X)_{t+1}$, the subspace can be recovered as $X_{t+1} = X - (E_X)_{t+1}$.

**2) *Given others, update the recovered clean soft labels:***
We optimize $E_F$ and $F$ by using the robust adaptive LP problem in Eq. (12). Based on the properties of the $l_{2,1}$-norm, $\| E_F^T \|_{2,1} = 2tr(E_F D E_F^T)$, where $D$ is a diagonal matrix with entries $d_{ii} = 1/(2 \| g^i \|_2)$ and $g^i$ is the $i$-th column vector of $E_F$. If each $g^i \neq 0$, where $i = 1,2,...,N$, the problem formulation in Eq. (12) can be re-expressed as

$$\min_{F, E_F, D} tr \left( (F - E_F) A (F - E_F)^T \right) + 2\alpha tr \left( E_F D E_F^T \right)$$
$$+ tr \left[ ((F - E_F) - Y) U ((F - E_F) - Y)^T \right] \quad , \quad (17)$$

where we also initialize the diagonal matrix $D$ as an identity matrix. Next, we fix $F$ to update $E_F$. By differentiating with respect to $E_F$ and setting it to zero, we obtain

$$(E_F)_{t+1} = (F_t A_t - YU + F_t U)(A_t + U + 2\alpha D_t)^{-1}. \quad (18)$$

After $(E_F)_{t+1}$ has been obtained, we can update $D_{t+1}$ as

$$D_{t+1} = diag\left( (d_{ii})_{t+1} \right), where \, (d_{ii})_{t+1} = 1/\left[ 2 \| g_{t+1}^i \|_2 \right], \, \forall i, \quad (19)$$

where $g_{t+1}^i$ is the $i$-th column of $(E_F)_{t+1}$. After the error term $(E_F)_{t+1}$ has been obtained, we update $F$ by differentiating with respect to $F$ and setting it to zero as follows:

$$F_{t+1} = \left( (E_F)_{t+1} A_t - (E_F)_{t+1} U + YU \right)(A_t + U)^{-1}. \quad (20)$$

After both $(E_F)_{t+1}$ and $F_{t+1}$ have been obtained, we can recover the underlying clean soft labels as $F_{t+1} = F_{t+1} - (E_F)_{t+1}$.

**3) *Given others, update the recovered clean weights:***
We compute $W$ and $E_W$ to recover the clean weight space via Eq. (11). Let $H = [F - E_F; X - E_X; e^T]$. By a similar argument, $\| E_W^T \|_{2,1} = 2tr(E_W O E_W^T)$, where $O$ is a diagonal matrix with $o_{ii} = 1/(2 \| \varphi^i \|_2)$ as entries and $\varphi^i$ is the $i$-th row of $E_W$. Assume that each $\varphi^i \neq 0$, Eq. (11) can be reformulated as

$$\min_{W, E_W, O} \left\| H - H(W - E_W) \right\|_F^2 + \left\| (W - E_W) \right\|_F^2 + \alpha \gamma \left\| E_W \right\|_{2,1}$$
$$= tr \left( (H - HW + HE_W)^T (H - HW + HE_W) \right) \quad , \quad (21)$$
$$+ tr \left( (W - E_W)^T (W - E_W) \right) + 2\alpha \gamma tr \left( E_W^T O E_W \right)$$

where $O$ is also initialized to an identity matrix. Then, we fix $W$ to update $E_W$. Differentiating the above equation with respect to $E_W$ yields

$$(E_W)_{t+1} = (H^T H + I + 2\alpha\gamma O)^{-1}(H^T H W_{t+1} + W_{t+1} - H^T H), \quad (22)$$

where $H_{t+1} = \left[ F_{t+1}^T - (E_F)_{t+1}^T; X - (E_X)_{t+1}^T; e \right]^T = \left[ \tilde{F}_{t+1}^T; \tilde{X}_{t+1}^T; e \right]^T$. After $(E_W)_{t+1}$ has been obtained, we can update matrix $O_{t+1}$ as

$$O_{t+1} = diag((o_{ii})_{t+1}), where (o_{ii})_{t+1} = 1 / \left[ 2 \|\varphi_{t+1}^i\|_2 \right], \forall i. \quad (23)$$

Similarly, we obtain the weights $W$ by differentiating the problem in Eq. (21) with respect to $W$ and setting it to zero:

$$W_{t+1} = (H_{t+1}^T H_{t+1} + I)^{-1}(H_{t+1}^T H_{t+1} + (E_W)_{t+1}), \quad (24)$$

$(W_{t+1})_{ii} = 0$ and $(W_{t+1}) = \max(W_{t+1}, 0)$, where $(W_{t+1})_{ii} = 0$ replaces the diagonal entries with zeros, and $(W_{t+1}) = \max(W_{t+1}, 0)$ is the constraint for replacing the negative entries with zeros to ensure the nonnegativity of $W_{t+1}$. After $(E_W)_{t+1}$ and $W_{t+1}$ have been obtained, we can recover the clean adaptive weight matrix as $\tilde{W}_{t+1} = W_{t+1} - (E_W)_{t+1}$.

Thus, by performing the classification and auto-weighted learning jointly in the recovered clean data, clean label and clean weighting spaces, the data representation and classification performances can be potentially improved. An early version of this paper has been presented in [56]. This paper also investigates the recovery of the clean weighting space and conducts thorough experimental evaluations on face, object, handwriting digit and text classification. To the best of our knowledge, joint recovery on the data, soft labels and adaptive weights has not been explored in previous research. For a complete presentation of the method, we summarize the optimization procedures of our ALP-TMR in Table I.

### D. Convergence Analysis

Since our ALP-TMR solves for the variables alternatively, we aim at analyzing its convergence behavior. To facilitate the proof, a lemma in [29] is reviewed.

**Lemma 1.** For any pair of non-zero vectors $p, q \in \mathbb{R}^m$, the following inequality holds:

$$\|p\|_2 - \frac{\|p\|_2^2}{2\|q\|_2} \leq \|q\|_2 - \frac{\|q\|_2^2}{2\|q\|_2}. \quad (25)$$

Thus, the convergence behavior of our ALP-TMR can be summarized by the following proposition.

**Proposition 1.** The objective function of ALP-TMR in Eq. (9) is non-increasing at each iteration, where the diagonal matrices $Q$, $D$ and $O$ are fixed as constants.

*Proof.* Denote the diagonal matrices $Q$, $D$ and $O$ at the *t*-th iteration as $Q_t$, $D_t$ and $O_t$, respectively. When we calculate $F_{t+1}$, $W_{t+1}$, $(E_F)_{t+1}$, $(E_X)_{t+1}$ and $(E_W)_{t+1}$ at the *(t+1)*-th iteration, the following inequality holds:

$$\mathbb{Q}(F_{t+1}, W_{t+1}, (E_F)_{t+1}, (E_X)_{t+1}, (E_W)_{t+1})$$
$$+ tr\left[ ((F_{t+1} - (E_F)_{t+1}) - Y)U((F_{t+1} - (E_F)_{t+1}) - Y)^T \right] + \alpha\Xi_{t+1}$$
$$\leq \mathbb{Q}(F_t, W_t, (E_F)_t, (E_X)_t, (E_W)_t)$$
$$+ tr\left[ ((F_t - (E_F)_t) - Y)U((F_t - (E_F)_t) - Y)^T \right] + \alpha\Xi_t$$
, (26)

where $\Xi_t$ is an auxiliary matrix, which is constructed as $\Xi_t = tr((E_F)_t D_t (E_F)_t^T) + tr(\beta(E_X)_t Q_t(E_X)_t^T) + tr(\gamma(E_W)_t^T O_t(E_W)_t)$. $\|(E_F)_{t+1}^T\|_{2,1} = \sum_i \|g_t^i\|_2$, $\|(E_X)_{t+1}^T\|_{2,1} = \sum_i \|b_t^i\|_2$ and $\|(E_W)_{t+1}\|_{2,1} = \sum_i \|\varphi_t^i\|_2$, where $i = 1, 2, ..., N$. Therefore, the following equivalent inequality holds:

$$\mathbb{Q}(F_{t+1}, W_{t+1}, (E_F)_{t+1}, (E_X)_{t+1}, (E_W)_{t+1})$$
$$+ tr\left[ ((F_{t+1} - (E_F)_{t+1}) - Y)U((F_{t+1} - (E_F)_{t+1}) - Y)^T \right]$$
$$+ \alpha \|(E_F)_{t+1}^T\|_{2,1} + \alpha \sum_{i=1}^N \left( \frac{\|g_{t+1}^i\|_2^2}{2\|g_t^i\|_2} - \|g_{t+1}^i\|_2 \right)$$
$$+ \alpha\beta \|(E_X)_{t+1}^T\|_{2,1} + \alpha\beta \sum_{i=1}^N \left( \frac{\|b_{t+1}^i\|_2^2}{2\|b_t^i\|_2} - \|b_{t+1}^i\|_2 \right)$$
$$+ \alpha\gamma \|(E_W)_{t+1}\|_{2,1} + \alpha\gamma \sum_{i=1}^N \left( \frac{\|\varphi_{t+1}^i\|_2^2}{2\|\varphi_t^i\|_2} - \|\varphi_{t+1}^i\|_2 \right)$$
$$\leq \mathbb{Q}(F_t, W_t, (E_F)_t, (E_X)_t, (E_W)_t)$$
$$+ tr\left[ ((F_t - (E_F)_t) - Y)U((F_t - (E_F)_t) - Y)^T \right]$$
$$+ \alpha \|(E_F)_t^T\|_{2,1} + \alpha \sum_{i=1}^N \left( \frac{\|g_t^i\|_2^2}{2\|g_t^i\|_2} - \|g_t^i\|_2 \right)$$
$$+ \alpha\beta \|(E_X)_t^T\|_{2,1} + \alpha\beta \sum_{i=1}^N \left( \frac{\|b_t^i\|_2^2}{2\|b_t^i\|_2} - \|b_t^i\|_2 \right)$$
$$+ \alpha\gamma \|(E_W)_t\|_{2,1} + \alpha\gamma \sum_{i=1}^N \left( \frac{\|\varphi_t^i\|_2^2}{2\|\varphi_t^i\|_2} - \|\varphi_t^i\|_2 \right)$$
. (27)

From the inequality in Lemma 1, it follows that

$$\frac{\|g_{t+1}^i\|_2^2}{2\|g_t^i\|_2} - \|g_{t+1}^i\|_2 \leq \frac{\|g_t^i\|_2^2}{2\|g_t^i\|_2} - \|g_t^i\|_2$$
$$\frac{\|b_{t+1}^i\|_2^2}{2\|b_t^i\|_2} - \|b_{t+1}^i\|_2 \leq \frac{\|b_t^i\|_2^2}{2\|b_t^i\|_2} - \|b_t^i\|_2 \quad . \quad (28)$$
$$\frac{\|\varphi_{t+1}^i\|_2^2}{2\|\varphi_t^i\|_2} - \|\varphi_{t+1}^i\|_2 \leq \frac{\|\varphi_t^i\|_2^2}{2\|\varphi_t^i\|_2} - \|\varphi_t^i\|_2$$

By combining the inequalities in Eq. (27) and Eq. (28), we obtain the following inequality:

$$\mathbb{Q}(F_{t+1}, W_{t+1}, (E_F)_{t+1}, (E_X)_{t+1}, (E_W)_{t+1})$$
$$+ tr\left[ ((F_{t+1} - (E_F)_{t+1}) - Y)U((F_{t+1} - (E_F)_{t+1}) - Y)^T \right]$$
$$+ \alpha\left( \|(E_F)_{t+1}^T\|_{2,1} + \beta\|(E_X)_{t+1}^T\|_{2,1} + \gamma\|(E_W)_{t+1}\|_{2,1} \right)$$
$$\leq \mathbb{Q}(F_t, W_t, (E_F)_t, (E_X)_t, (E_W)_t)$$
$$+ tr\left[ ((F_t - (E_F)_t) - Y)U((F_t - (E_F)_t) - Y)^T \right]$$
$$+ \alpha\left( \|(E_F)_t^T\|_{2,1} + \beta\|(E_X)_t^T\|_{2,1} + \gamma\|(E_W)_t\|_{2,1} \right)$$
, (29)

which indicates that the objective function value of our ALP-TMR will monotonically decrease in the iterations. Besides, since the objective function has a lower bound, namely, zero, the above optimization converges. Although the above proposition can indicate that the objective function of ALP-TMR is non-increasing, we still hope that the major variable $F$ can also converge. Therefore, the convergence condition is simply set as $Error(t+1) = \|F_{t+1} - F_t\|_F \leq 10^{-3}$, which ensures that the prediction result will not change drastically.

TABLE I.
OPTIMIZATION PROCEDURE OF OUR ALP-TMR.

**Input:** Training data matrix $X$, initial label matrix $Y$, parameters $\alpha$, $\beta$ and $\gamma$.
Initialize $Q$, $D$ and $O$ to the identity matrices; Initialize the weight matrix $W$ by the reconstruction weights [10]; Initialize the error matrices $E_F$, $E_W$ and $E_X$ to zero matrices; $F_0 = Y$; $t = 0$; $u_i^l = 1$, $u_i^u = 0$.
*while* not converged *do*
1: $t \leftarrow t+1$;
2: Recover the error matrix $(E_X)_{t+1}$ via Eq. (15) and obtain the recovered clean data as $X - (E_X)_{t+1}$;
3: Estimate the soft label matrix $F_{t+1}$ via Eq. (20);
4: Recover the error matrix $(E_F)_{t+1}$ via Eq. (18) and recover the clean soft labels as $F_{t+1} - (E_F)_{t+1}$;
5: Update the weight matrix $W_{t+1}$ via Eq. (24);
6: Recover the error $(E_W)_{t+1}$ via Eq. (22) and obtain the recovered clean weights as $W_{t+1} - (E_W)_{t+1}$;
7: Update the diagonal matrices $Q_{t+1}$, $D_{t+1}$, and $O_{t+1}$ by using Eq. (16), Eq. (19) and Eq. (23), respectively;
*end while*
**Output:** Recovered clean soft labels $F^* = F_{t+1}$ and clean adaptive weight matrix $W^* = W_{t+1}$.

## IV. SIMULATION RESULTS AND ANALYSIS

We evaluate ALP-TMR mainly on data classification and compare the results with those of 9 related popular methods: GFHF [1], LLGC [5], SLP [3], LNP [2], CD-LNP [4], ProjLP [6], PN-LP [7], SparseNP [8] and AdaptiveNP [9]. Since all the compared algorithms must pre-obtain a similarity graph weight matrix for measuring the manifold smoothness during the label propagation process, and to avoid the complicated issue of choosing the optimal kernel width the weighting step that was encountered in previous work, we compute the same reconstruction weights [10] for each compared method for fair comparison. The nearest-neighbor search involves another troublesome parameter, namely, the number of nearest neighbors K, which is also difficult to determine in practice. In this study, K is set to 7 for each method according to [33], in which it is shown that this choice performs well in most cases. We also symmetrize the reconstruction weights and set the diagonal entries to zeros for LLGC. For fair comparison, the model parameters of each algorithm are all carefully chosen via grid search. We conduct all experiments on a PC with Intel (R) Core (TM) i5-4590 @ 3.30 Hz and 8.00 GB.

TABLE II.
DESCRIPTION OF THE REAL-WORLD DATASETS

| Dataset Name | #Classes (c) | #Dim (n) | #Points |
|---|---|---|---|
| YaleB-UMIST [11][27] | 58 | 1024 | 3426 |
| CMU PIE [12] | 68 | 1024 | 11554 |
| USPS [13] | 10 | 256 | 11000 |
| CASIA-HWDB1.1 [22] | 10 | 196 | 2381 |
| ETH80 [16] | 80 | 1024 | 3280 |
| COIL100 [15] | 100 | 1024 | 7200 |
| TDT2[1] | 30 | 2000 | 9394 |
| RCV1 [51] | 4 | 2000 | 9625 |
| CIFAR100[2] | 100 | 1600 | 50000 |

In this simulation, nine public image and text databases are employed, namely, two face image databases, three object image databases, two handwriting digit databases and two document datasets. TABLE II presents detailed information on the considered datasets. As is common practice, each face/object image is resized to 32×32 pixels; hence, each image corresponds to a data point in a 1024-dimensional space. For the transductive classification on each database, each database is split into a labeled set and an unlabeled set, similar to [1-9][21]. Finally, the accuracy is computed by comparing the predicted labels of the unlabeled samples with the ground-truth labels that are provided by the original data corpus. To avoid bias, we average all the results over 10 random splits of labeled/unlabeled data points under each setting.

### A. Parameter Sensitivity Analysis

The values of the model parameters ($\alpha$, $\beta$, and $\gamma$) in the objective function of our ALP-TMR may affect the performance; hence, we explore the effects of the parameters on the classification results. We present the classification results of ALP-TMR under various parameters on the YaleB-UMIST database as an example. YaleB-UMIST is a mixed face database of extended YaleB [11] and UMIST [27]. The extended YaleB face database is highly challenging because it contains images with various facial expressions and illumination conditions. The UMIST face database contains 1012 images of 20 persons of mixed race/gender/appearance. For this study, we select 40 samples per subject as labeled and test on the remaining samples. Since ALP-TMR has three parameters, we fix one parameter and tune other two via the grid search strategy. First, we fix $\alpha = 10^{-8}$ to tune $\beta$ and $\gamma$. Then, we fix $\beta = 10^6$ to tune $\alpha$ and $\gamma$. Finally, we fix $\gamma = 10^{-2}$ to tune $\alpha$ and $\beta$. The classification results are averaged over 30 runs in each setting, and we tune $\alpha$, $\beta$ and $\gamma$ in Fig. 1, where the parameters are selected from the candidate set $\{10^{-8}, 10^{-6}, \ldots, 10^8\}$. According to Fig. 1, ALP-TMR performs effectively for a wide range of parameter settings, namely, our ALP-TMR is robust to the model parameters.

In addition the above parameter analysis, we explore the effects of the model parameters in the objective function of our ALP-TMR by setting $\alpha = 0$, $\beta = 0$ and $\gamma = 0$ alternatively. $\alpha = 0$ indicates that the sparse errors in the original data $X$, soft labels $F$ and adaptive weights $W$ cannot be modeled and removed. If $\beta = 0$, the sparse errors in data $X$ cannot be modeled. If $\gamma = 0$, the sparse errors in the weights $W$ cannot be modeled. In this simulation, four real-world image databases, namely, CMU PIE, YaleB, ETH80 and COIL100, are adopted. For transductive classification, the numbers of labeled data points are set to 12, 30, 100 and 15 for CMU PIE, YaleB, ETH80 and COIL100, respectively, and the remaining samples are set as unlabeled. The classification results that are obtained by setting $\alpha = 0$, $\beta = 0$ and $\gamma = 0$ over CMU PIE, YaleB, ETH80 and COIL100 are presented in Table III, from which we find that setting $\alpha = 0$, $\beta = 0$ or $\gamma = 0$ strongly degraded the classification results, namely, the involved terms that are associated with $\alpha$, $\beta$ and $\gamma$ are all important for improving the results of our method.

TABLE III.
CLASSIFICATION RESULTS OVER VARIOUS PARAMETERS ON CMU PIE, YaleB, ETH80 AND COIL100.

| Settings of ALP-TMR | CMU PIE | YaleB | ETH80 | COIL100 |
|---|---|---|---|---|
| $\alpha = 0, \beta \neq 0, \gamma \neq 0$ | 0.9034 | 0.7933 | 0.7459 | 0.8249 |
| $\alpha \neq 0, \beta = 0, \gamma \neq 0$ | 0.8889 | 0.7570 | 0.7293 | 0.8030 |
| $\alpha \neq 0, \beta \neq 0, \gamma = 0$ | 0.9139 | 0.8194 | 0.7838 | 0.8541 |
| $\alpha \neq 0, \beta \neq 0, \gamma \neq 0$ | **0.9562** | **0.9239** | **0.8486** | **0.8991** |

---

[1] https://www.nist.gov/speech/tests/tdt/tdt98/index.htm

[2] http://www.cs.toronto.edu/~kriz/cifar.html

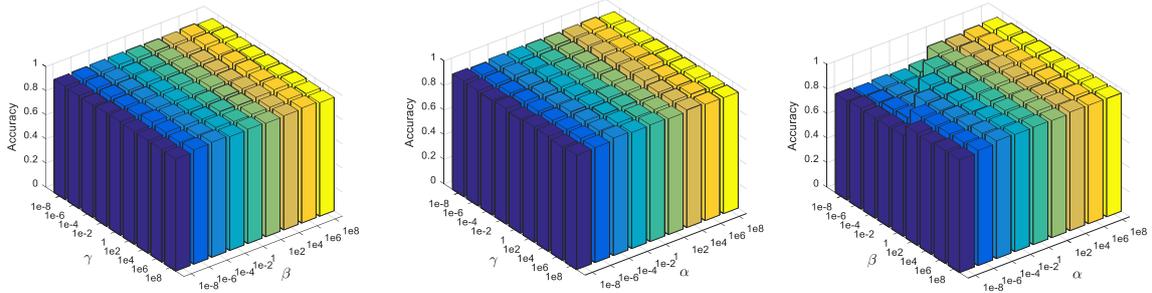

**Fig. 1:** Results under various parameters: (left) fix $\gamma$ to tune $\alpha$ and $\beta$ ; (middle) fix $\alpha$ to tune $\beta$ and $\gamma$ ; and (right) fix $\beta$ to tune $\alpha$ and $\gamma$ .

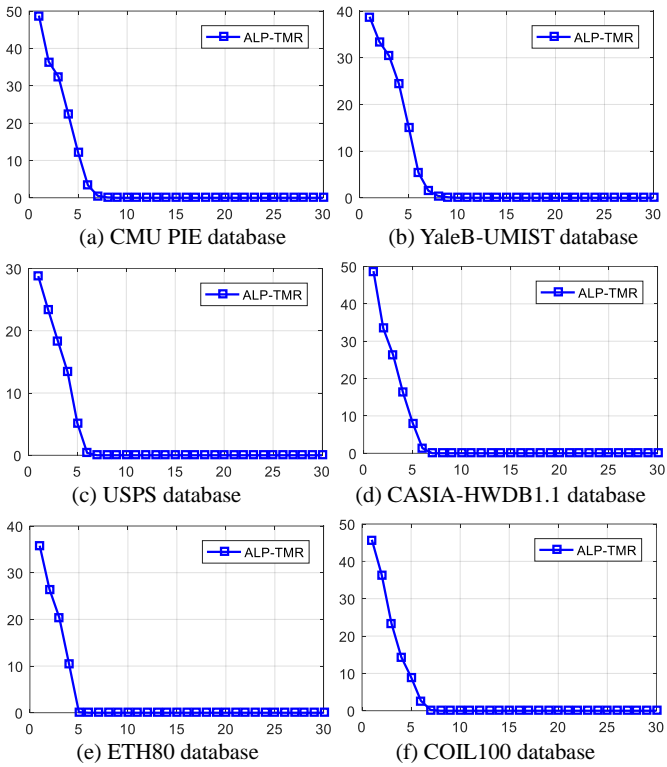

(a) CMU PIE database
(b) YaleB-UMIST database
(c) USPS database
(d) CASIA-HWDB1.1 database
(e) ETH80 database
(f) COIL100 database

**Fig. 2:** Convergence results of our proposed ALP-TMR method on six real image databases.

### B. Convergence Analysis

We present numerical results and use them to evaluate the convergence of ALP-TMR in this study. Two real face databases, two handwriting digit databases and two object databases, namely, YaleB-UMIST, CMU PIE [12], USPS [13], CASIA-HWDB1.1 [22], ETH80 [16], and COIL100 [15], are applied to evaluate our method. For each database, we fix the number of labeled samples, namely, to 10 for CMU PIE, 30 for YaleB-UMIST, 100 for USPS, 40 for CASIA-HWDB1.1, 100 for ETH80 and 30 for COIL100, to evaluate the differences between consecutive estimated label matrices $F$. The convergence analysis results of our ALP-TMR on the six image databases are presented in Fig. 2. The divergence between the consecutive soft label matrices $F$ decreases during the iterations and finally converges to zero; hence, the final result will not differ drastically. In addition, ALP-TMR typically converges in less than 10 iterations, namely, the convergence of ALP-TMR is fast in the investigated cases.

### C. Exploratory Data Analysis via Visualization

We visualize the recovered estimated soft labels and the recovered weights for visual evaluations. The CMU PIE face image database is evaluated in this simulation.

**Visualization of the recovered clean predicted soft labels.** First, we visualize the predicted soft labels. For GFHF, LLGC, LNP, SparseNP, AdaptiveNP and our ALP-TMR, we aim at visualizing the estimated soft labels for the unlabeled images in Fig. 3, where the horizontal axis in each subfigure corresponds to the number of unlabeled images and the vertical axis to the number of classes. We select 20 face images from each person as the labeled set and 22 images as unlabeled samples for transductive learning. The brightness in each gray figure is determined by the values in the estimated soft labels, namely, the corresponding pixels will be brighter if the reconstruction value is larger. We also observe that the face images of the same subject are effectively merged into clusters in the results of ALP-TMR, namely, fewer mixed signs are included in the estimated results of our ALP-TMR due to the positive effects of robust soft label recovery.

In addition, to evaluate the effects of the robust soft label recovery, namely, $F = \tilde{F} + E_F$ , we conduct a simulation for visualizing the decomposition results in Fig. 4, where we present the original estimated label matrix $F$, the recovered clean soft label matrix $\tilde{F}$ and the recovered sparse error $E_F$ . The recovered soft label matrix $\tilde{F}$ contains less noise and fewer unfavorable mixed signs than the original label matrix $F$; hence, the label recovery can indeed remove the unfavorable mixed signs and noise. Moreover, estimating the hard labels of the unlabeled samples based on the recovered clean labels in $\tilde{F}$ will be potentially more accurate than based on the original soft label matrix $F$.

**Visualization of the recovered clean adaptive weights.** For accurate representation and similarity measures, the weight matrix should exhibit a powerful descriptive performance based on various classes and, moreover, should have a block-diagonal structure so that one can reconstruct each sample from the samples of one class as much as possible, which could improve the subsequent data classification result. Thus, we also prepare a study for evaluating the robust weight recovery, namely, $W = \tilde{W} + E_W$ , by visualizing the weight matrix that is recovered by ALP-TMR in Fig. 5, where we present the original weight matrix $W$, the recovered clean weights $\tilde{W}$ and the recovered sparse error $E_W$ . For this study, we select 15 images from each person as labeled. From the visualization results, we observe the following: (1) both the recovered weights $\tilde{W}$ and the original weights $W$ have block-diagonal structures, but the structure of the recovered weights $\tilde{W}$ is clearer; (2) the intra-class connectivity of the recovered weights $\tilde{W}$ is similar to that of $W$, namely, the weight recovery process can effectively preserve the important intra-class connectivity; and (3) the recovered weight matrix $\tilde{W}$ contains fewer incorrect inter-class connections than $W$; hence, the robust weight recovery can remove the incorrect inter-class connections effectively. Thus, the encoded manifold structures and pairwise similarities that are obtained using the recovered weights $\tilde{W}$ will be more accurate, which will improve the label estimation results.

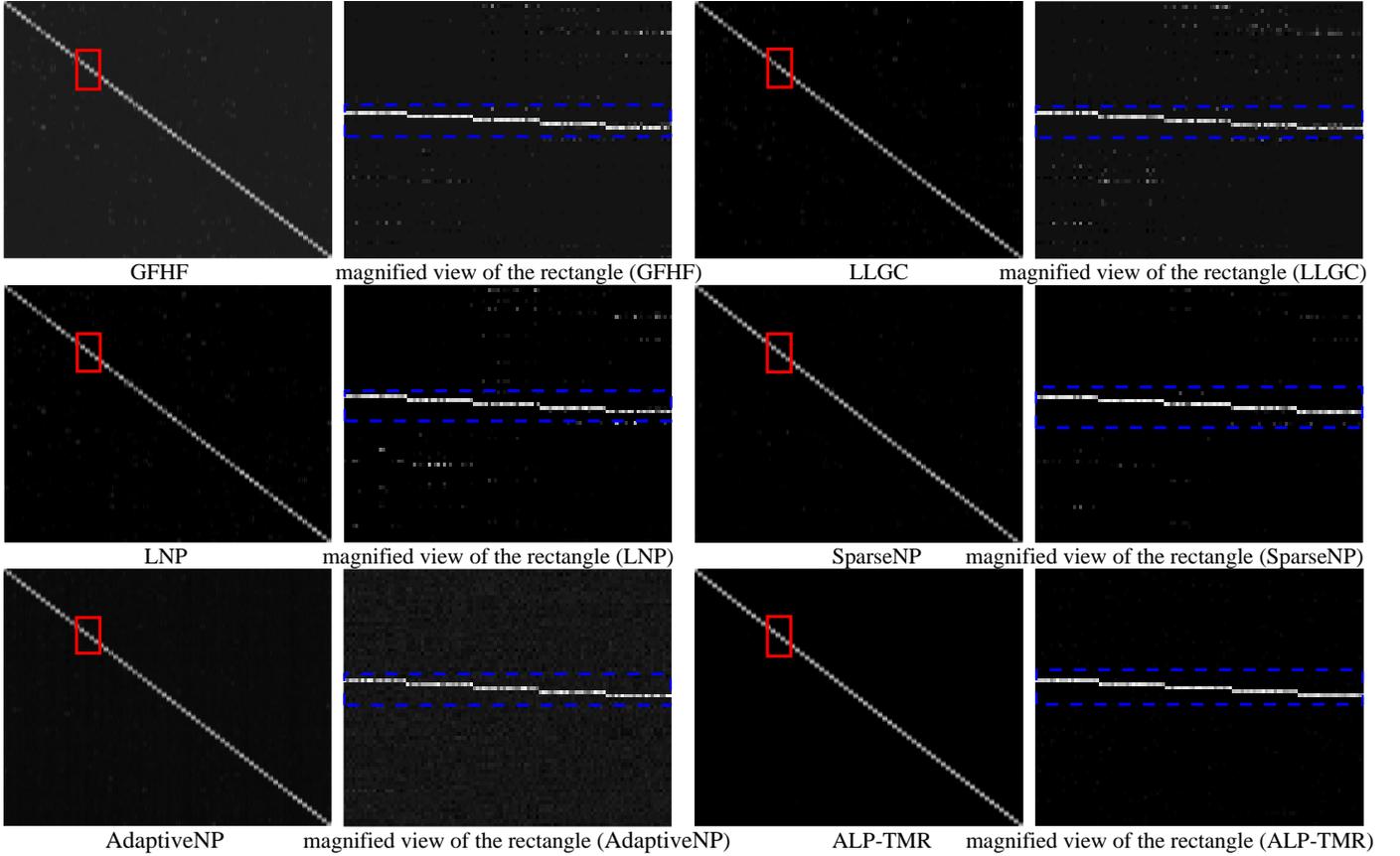

**Fig. 3:** Visualization of the predicted soft label matrix of each semi-supervised algorithm on the CMU PIE face image database.

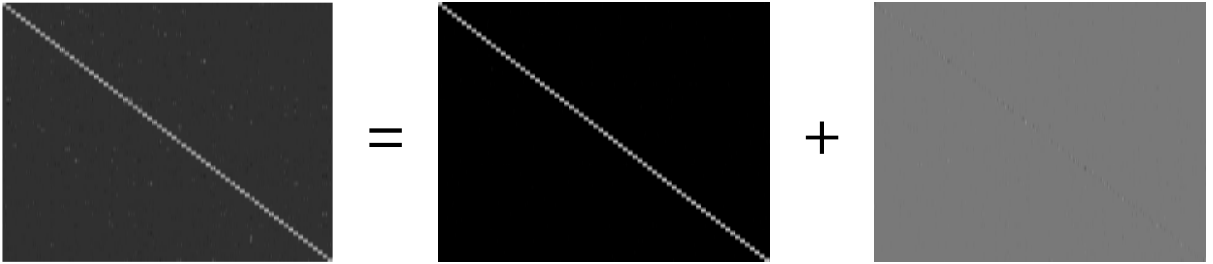

**Fig. 4:** Visualization of the soft label recovery via $F = \tilde{F} + E_F$ on the CMU PIE face image database: (left) original soft labels $F$, (middle) recovered clean soft labels $\tilde{F}$ and (right) recovered sparse error $E_F$.

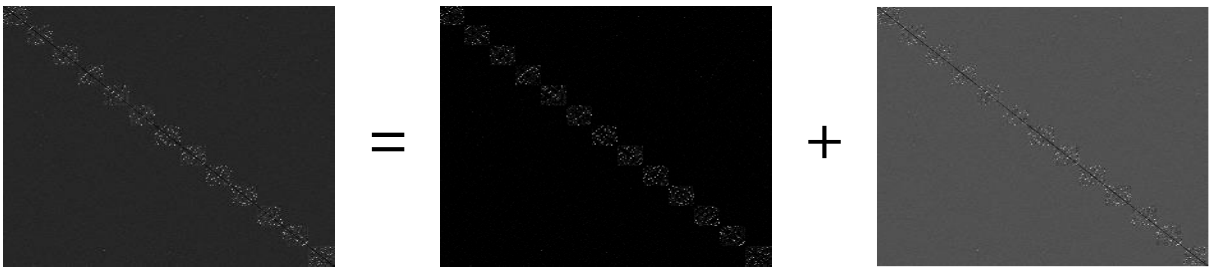

**Fig. 5:** Visualization of the decomposition $W = \tilde{W} + E_W$ on the CMU PIE face database: (left) original weight matrix $W$, (middle) recovered clean weight matrix $\tilde{W}$ and (right) recovered sparse error $E_W$.

### D. Face Recognition

We evaluate each algorithm on face image recognition and representation on two face datasets, namely, YaleB-UMIST and CMIU PIE. Sample images from these databases are shown in Fig. 6. We compute the classification result of each algorithm by changing the number of labeled samples in the training set and averaging the classification results over 10 random splits of labeled and unlabeled images. In this paper, the random feature extractor [24] is employed for each face database, and each face image is projected onto a 500-D feature vector. We compare the classification results of our ALP-TMR with those of the SLP, GFHF, LNP, LLGC, CD-LNP, ProjLP, SparseNP, AdaptiveNP and PN-LP methods.

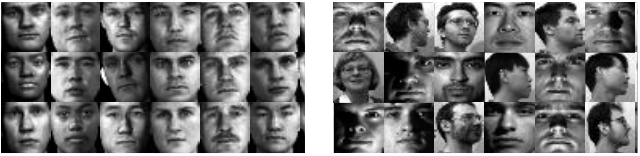

**Fig. 6:** Image samples from YaleB-UMIST (left) and CMU PIE (right).

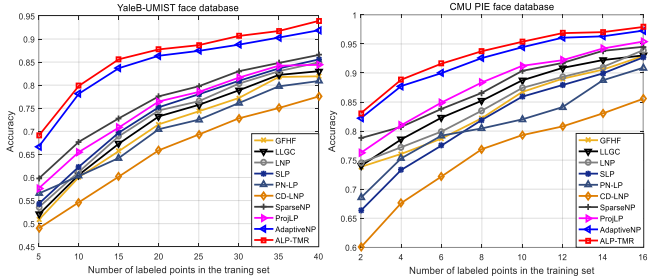

**Fig. 7:** Classification results of each algorithm with various numbers of labeled data on YaleB-UMIST (left) and CMU PIE (right).

TABLE III:
PERFORMANCE COMPARISON OF RANDOM FEATURES ON TWO FACE DATABASES.

| Data Method | YaleB-UMIST Mean±STD | Best (%) | CMU PIE Mean±STD | Best (%) |
|---|---|---|---|---|
| GFHF | 71.11±9.76 | 82.92 | 83.73±7.05 | 92.92 |
| LLGC | 71.75±9.09 | 83.01 | 85.25±7.06 | 93.02 |
| LNP | 72.49±9.82 | 84.97 | 84.61±6.94 | 93.92 |
| SLP | 73.56±9.81 | 84.67 | 82.19±9.18 | 92.67 |
| PN-LP | 69.16±8.59 | 79.89 | 81.16±7.16 | 90.89 |
| CD-LNP | 66.43±9.24 | 77.55 | 76.30±8.97 | 85.55 |
| SparseNP | 75.83±8.83 | 85.50 | 87.45±6.12 | 94.50 |
| ProjLP | 73.95±8.94 | 82.41 | 87.33±6.60 | 95.41 |
| AdaptiveNP | 85.04±9.58 | 93.89 | 92.44±5.51 | 96.79 |
| **ALP-TMR** | **86.45±7.89** | **94.92** | **93.57±5.44** | **97.64** |

**Recognition on the YaleB-UMIST database.** We evaluate each algorithm for face representation and recognition on the mixed YaleB-UMIST face database. The classification results of each method under various numbers of labeled samples in the training set are presented in Fig. 7 (left), where we report the classification accuracy based on the unlabeled set. According to Fig. 7, the averaged results (%) with the standard deviation (STD) are listed in TABLE III. The best records over various random splits are also specified. The number of labeled images per subject is set to 5, 10, 15, 20, 25, 30, 35, 40 and 45. We find that (1) the classification result of each model is improved by increasing the number of labeled images and (2) our ALP-TMR outperforms the compared methods by obtaining better results in most cases. The recently proposed AdaptiveNP also performs well on this face database by delivering comparable results to ALP-TMR. SparseNP and ProjLP deliver better results than the remaining methods since they introduce two feasible approaches for reducing the noise and unfavorable mixed signs from the estimated soft labels via label embedding and sparse constraint, respectively. CD-LNP performs the worst on this dataset.

**Recognition on the CMU PIE face database.** This database contains face images under various lighting and illumination conditions and facial expressions. The performances of each method with various numbers of labeled images are presented in the right side of Fig. 7. For this database, we choose 2, 4, 6, 8, 10, 12, 14 and 16 images from each person as labeled and use the remaining images for testing. TABLE III lists the mean accuracy (%) and highest accuracy (%) according to the results in Fig. 7. We find that (1) ALP-TMR obtains the enhanced results compared with the related algorithms, which can be attributed to the strategy of triple matrix recovery; and (2) AdaptiveNP obtains highly competitive results with ALP-TMR. ProjLP realizes higher accuracies than other methods, and CD-LNP performs the worst.

### E. Handwriting Digit Recognition

We evaluate each method on recognizing the handwritten digits on the USPS [13] and CASIA-HWDB1.1 [22] databases. Examples of handwritten digits are shown in Fig. 8.

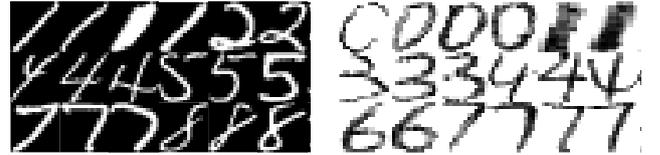

**Fig. 8:** Image samples from USPS (left) and CASIA-HWDB1.1 (right).

**Results on the USPS database.** This handwriting digit database (available from http://www.cs.toronto.edu/~roweis/data.html) contains images of digits from '0' to '9' of size 16×16 pixels, and each digit corresponds to 1100 images, namely, each image corresponds to a 256-D vector. The classification results of each method with various numbers of labeled images are presented in Fig. 9. The averaged results according to Fig. 9 are listed in Table IV. In this study, the number of labeled digit images is fixed to 5, 10, …, 60. We find that (1) the results of each criterion are improved as the number of labeled digits is increased and (2) ALP-TMR outperforms the other algorithms.

**Results on the CASIA-HWDB1.1 database.** In this study, a sampled subset, namely, HWDB1.1-D [43], which contains 2381 handwritten digits ('0'-'9'), from CASIA-HWDB1.1[14] is evaluated in this study. Following [43], we resize all images to 14×14 pixels for consistency of the digit sizes. The classification results that are delivered by each algorithm with various numbers of labeled digits in the training set are presented in Fig. 9, where the number of labeled digits is set to 2, 5, …, 35. The averaged results according to Fig. 9 are listed in Table IV. We find that ALP-TMR realizes higher accuracies than its competitors on this dataset. In addition, the label prediction results of each method can be improved by increasing the number of labeled images.

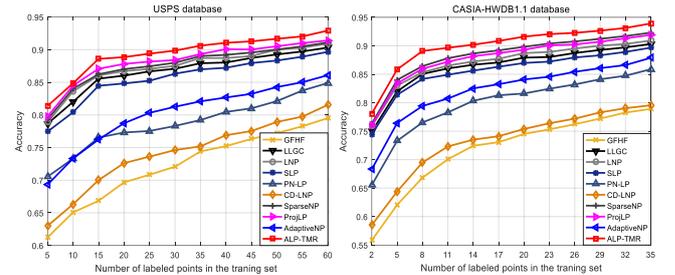

**Fig. 9:** Classification results of each algorithm with various numbers of labeled samples on USPS (left) and CASIA-HWDB1.1 (right).

TABLE IV:
PERFORMANCE COMPARISON OF ALGORITHMS ON TWO HANDWRITING DIGITS DATABASES.

| Data Method | USPS database Mean±STD | Best (%) | CASIA-HWDB1.1 Mean±STD | Best (%) |
|---|---|---|---|---|
| GFHF | 72.57±8.72 | 79.36 | 73.61±9.69 | 92.79 |
| LLGC | 85.34±6.52 | 89.95 | 85.97±2.65 | 90.74 |
| LNP | 86.57±6.25 | 89.72 | 86.41±2.21 | 90.83 |
| SLP | 85.89±5.25 | 88.67 | 85.61±2.62 | 90.12 |
| PN-LP | 80.23±6.24 | 85.63 | 81.54±6.87 | 79.75 |
| CD-LNP | 73.47±9.22 | 81.63 | 74.86±7.88 | 78.37 |
| SparseNP | 89.93±2.42 | 91.73 | 89.70±2.22 | 92.45 |
| ProjLP | 89.68±3.22 | 91.83 | 88.54±2.25 | 92.51 |
| AdaptiveNP | 82.37±5.99 | 86.73 | 85.67±4.03 | 88.03 |
| **ALP-TMR** | **91.12±3.04** | **93.82** | **90.76±2.63** | **93.92** |

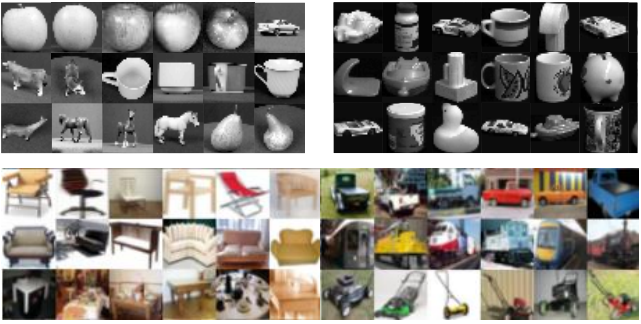

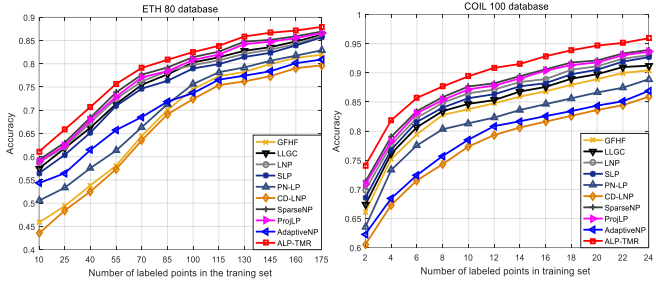

**Fig. 10:** Sample images of ETH80 (top left), COIL100 (top right) and CIFAR 100 (bottom row).

**Fig. 11:** Classification results of each algorithm with various numbers of labeled samples on ETH80 (left) and COIL100 (right).

TABLE V:
PERFORMANCE COMPARISON OF RANDOM FEATURES ON TWO OBJECT DATABASES.

| Data Method | ETH80 database Mean±STD | Best (%) | COIL100 database Mean±STD | Best (%) |
|---|---|---|---|---|
| GFHF | 71.86±5.50 | 81.85 | 86.36±3.78 | 90.92 |
| LLGC | 81.62±2.71 | 86.02 | 86.97±3.49 | 91.42 |
| LNP | 80.86±2.68 | 85.62 | 87.36±3.38 | 92.32 |
| SLP | 80.30±2.89 | 85.17 | 87.18±3.83 | 92.07 |
| PN-LP | 72.41±4.42 | 82.89 | 82.91±4.81 | 88.89 |
| CD-LNP | 70.11±4.83 | 79.55 | 76.36±5.34 | 85.55 |
| SparseNP | 82.81±2.02 | 88.30 | 90.41±3.10 | 93.50 |
| ProjLP | 82.39±1.95 | 87.61 | 90.14±3.47 | 93.01 |
| AdaptiveNP | 73.29±4.34 | 80.29 | 78.42±4.54 | 86.29 |
| **ALP-TMR** | **82.26±1.95** | **88.92** | **91.51±4.13** | **96.92** |

### B. Object Recognition

We evaluate each algorithm on recognizing the objects in three databases, namely, ETH80, COIL100 and CIFAR100. Examples from the two databases are shown in Fig. 10. We also use the random feature descriptor, and the dimension of the extracted random-object features is 400. For CIFAR100, we use principal component analysis (PCA) [50] to extract the principal features and to reduce the dimension to 1600.

**Results on ETH80.** This database has 8 large categories. Each large category contains 10 subcategories and each subcategory contains 41 images from various viewpoints. The classification results of each algorithm with various numbers of labeled data points in the training set are presented in Fig. 11. The statistics for Fig. 11 are listed in Table V. We find that (1) the performance of each algorithm is enhanced as the number of labeled images increases and (2) ALP-TMR obtains superior results compared to the other methods.

**Results on COIL100.** The object images of COIL100 were placed on a motorized turntable against a black background. The turntable was rotated 360 degrees to vary the poses with respect to the axed color camera. Images were captured at pose intervals of 5 degrees, which corresponds to 72 poses per object. Fig. 11 presents the classification results of each algorithm with various numbers of training images, and TABLE V lists the statistics for Fig. 11. We find that our proposed ALP-TMR delivers comparable or better results than the compared methods.

**Results on CIFAR100.** This object database includes 100 classes, each class has 600 images and the 100 classes are clustered into 20 super-classes. In this database, each image has a "fine" label, which specifies the class to which it belongs, and a "coarse" label, which specifies the superclass to which it belongs. The whole database is divided into a training set of 50,000 samples and a test set of 10,000 samples. We use the "fine" label for each model and perform transductive leaning based on the training set for efficiency. The classification results for each criterion under various numbers of labeled data points in the training set are listed in TABLE VI. We observe that (1) the classification result for each criterion is improved when the number of labeled images increases and (2) our ALP-TMR obtains enhanced results compared with the other models.

TABLE VI:
RECOGNITION PERFORMANCE COMPARISON OF PRINCIPAL FEATURES ON THE CIFAR100 DATABASE.

| Setting Method | CIFAR100 (50 labeled) Mean±STD | Best (%) | CIFAR100 (100 labeled) Mean±STD | Best (%) | CIFAR100 (150 labeled) Mean±STD | Best (%) | CIFAR100 (200 labeled) Mean±STD | Best (%) |
|---|---|---|---|---|---|---|---|---|
| GFHF | 58.63±0.83 | 59.53 | 64.83±0.72 | 65.23 | 71.03±1.02 | 72.17 | 79.87±1.38 | 80.83 |
| LLGC | 59.31±1.25 | 60.32 | 65.63±1.03 | 66.37 | 72.38±0.83 | 73.67 | 81.23±0.92 | 82.39 |
| LNP | 59.18±1.27 | 60.26 | 65.12±0.82 | 65.97 | 72.02±0.74 | 73.37 | 80.67±0.45 | 81.27 |
| SLP | 57.82±1.86 | 59.47 | 63.39±1.34 | 64.76 | 70.19±1.48 | 71.89 | 78.38±0.73 | 79.65 |
| PN-LP | 53.38±1.38 | 54.32 | 59.48±1.30 | 60.81 | 65.28±1.22 | 67.03 | 72.89±0.81 | 73.66 |
| CD-LNP | 52.08±1.40 | 53.03 | 58.30±1.27 | 59.93 | 64.48±0.88 | 65.74 | 71.83±0.92 | 73.36 |
| SparseNP | 60.47±1.34 | 61.92 | 67.28±1.48 | 68.77 | 73.98±1.20 | 74.29 | 82.37±0.83 | 83.83 |
| ProjLP | 60.23±1.27 | 61.33 | 66.78±1.03 | 67.44 | 74.91±0.75 | 75.47 | 82.34±0.58 | 83.92 |
| AdaptiveNP | 61.29±0.92 | 62.27 | 67.36±1.11 | 68.78 | 74.47±0.79 | 75.33 | 82.88±0.69 | 83.98 |
| **ALP-MD** | **62.17±1.39** | **64.62** | **68.67±1.30** | **69.17** | **75.33±0.69** | **76.47** | **83.23±0.48** | **84.86** |

### C. Document Categorization

In addition to examining each method for representing and recognizing images, we also conduct a study in which we evaluate them on document categorization on the two text databases: TDT2 and RCV1. For efficiency, PCA is also employed to reduce the numbers of dimensions of TDT2 and

RCV1 to 2000 for document categorization.

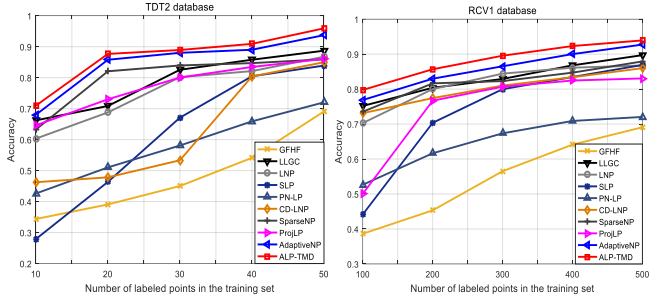

**Fig. 12:** Classification performances of each method with various numbers of labeled samples on TDT2 (left) and RCV1 (right).

TABLE VII:
PERFORMANCE COMPARISON OF ALGORITHMS ON TWO TEXT DATABASES.

| Data Method | RCV1 database Mean±STD | Best (%) | TDT2 database Mean±STD | Best (%) |
|---|---|---|---|---|
| GFHF | 54.74±12.7 | 69.08 | 48.33±13.7 | 69.08 |
| LLGC | 82.99±5.64 | 88.70 | 78.82±9.78 | 88.70 |
| LNP | 81.52±6.81 | 86.74 | 75.62±10.8 | 86.74 |
| SLP | 72.97±15.3 | 83.90 | 61.14±20.7 | 83.90 |
| PN-LP | 64.91±7.96 | 72.04 | 57.93±11.6 | 72.04 |
| CD-LNP | 80.18±5.05 | 84.98 | 62.55±17.6 | 84.98 |
| SparseNP | 82.01±5.34 | 85.91 | 79.97±9.54 | 85.91 |
| ProjLP | 74.59±12.3 | 86.05 | 77.44±8.67 | 86.83 |
| AdaptiveNP | 85.82±6.20 | 93.72 | 84.91±9.88 | 93.72 |
| **ALP-TMR** | **88.24±5.68** | **95.92** | **86.88±9.44** | **95.92** |

**Results on TDT2.** The TDT2 text database contains data that were collected during the first half of 1998 and obtained from 6 sources, namely, 2 newswires (APW and NYT), 2 radio programs (VOA and PRI) and 2 television programs (CNN and ABC). The TDT2 text database contains 11201 on-topic documents from 96 semantic categories. In this study, a sampled set of the original TDT2 corpus is applied [54], where documents that appear in two or more categories were removed and the largest 30 categories were kept, thereby leaving us with a total of 9,394 documents. The performances of each method under various training set sizes are presented in Fig. 12 and TABLE VII. The categorization performance is enhanced as the number of labeled images increases. Our proposed ALP-TMR obtains enhanced results compared to the other methods in most cases.

**Results on RCV1.** The RCV1 database contains information on topics, regions, and industries for each document and a hierarchical structure for topics and industries. In this study, a subset of 9625 documents with 29992 distinct words is used [55], which includes categories "C15", "ECAT", "GCAT" and "MCAT", which contain 2022, 2064, 2901, and 2638 documents, respectively. The classification results are presented in Fig. 12 and TABLE VII presents the statistics for Fig. 12. According to the results, ALP-TMR outperforms the compared methods. GFHF performs the worst on this set.

### D. Robustness Analysis

In this study, we evaluate the robustness of each semi-supervised algorithm against pixel corruptions. Two face databases (YaleB-UMIST and CMU PIE), two handwritten digit databases (CASIA-HWDB1.1 and USPS) and two object databases (ETH80 and COIL100) are evaluated. The gray values of the labeled image data are corrupted by including random Gaussian noise with a fixed variance. For noisy face recognition, we randomly select 40 and 20 images per person from the YaleB-UMIST and CMU PIE databases, respectively, as the labeled set. For the noisy handwritten digit recognition, we randomly select 50 and 20 samples from each digit of the USPS and CASIA-HWDB1.1 datasets, respectively, as the labeled set. For the noisy object recognition, we randomly select 70 and 30 sample images from each object class of the ETH80 and COIL100 databases, respectively, as the labeled set.

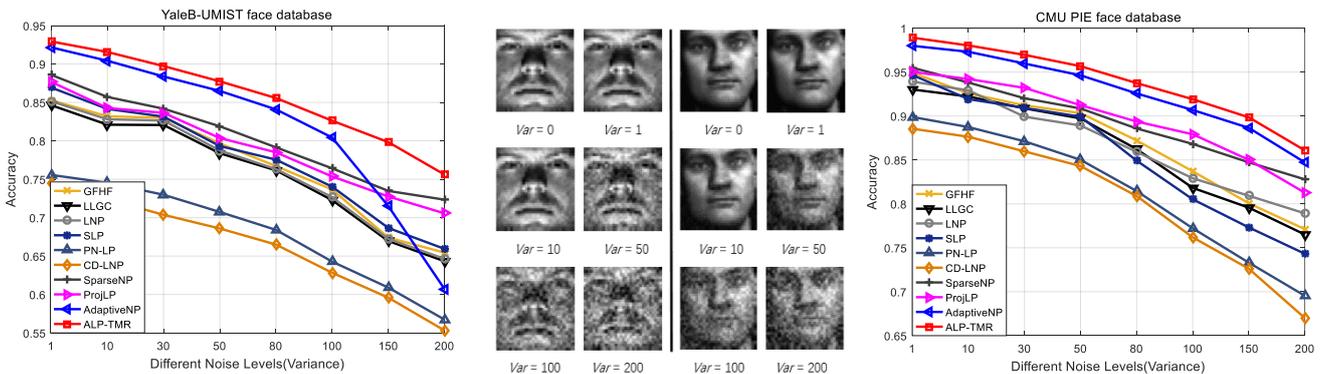

**Fig. 13:** Classification performances of each method with various levels of pixel corruption on YaleB-UMIST (left) and CMU PIE (right).

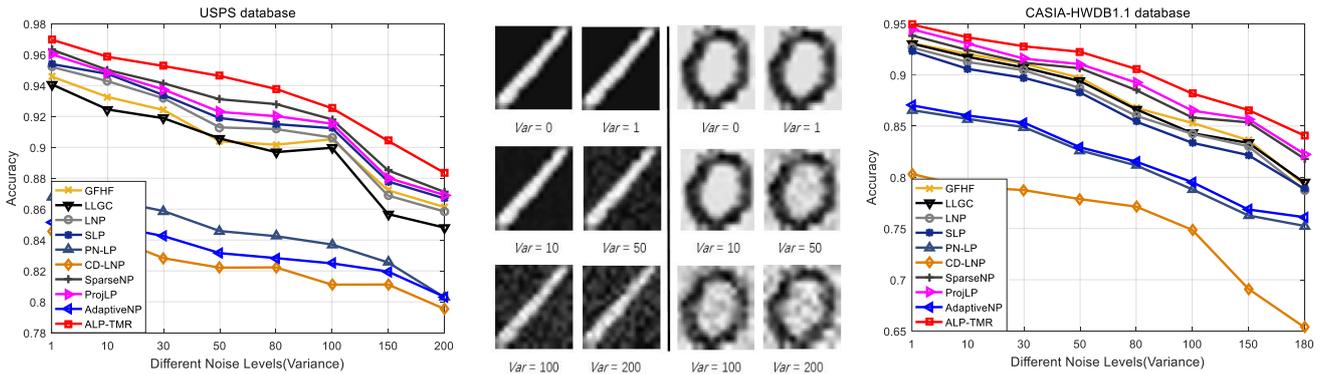

**Fig. 14:** Classification results of each algorithm under various levels of pixel corruption on USPS (left) and CASIA-HWDB1.1 (right).

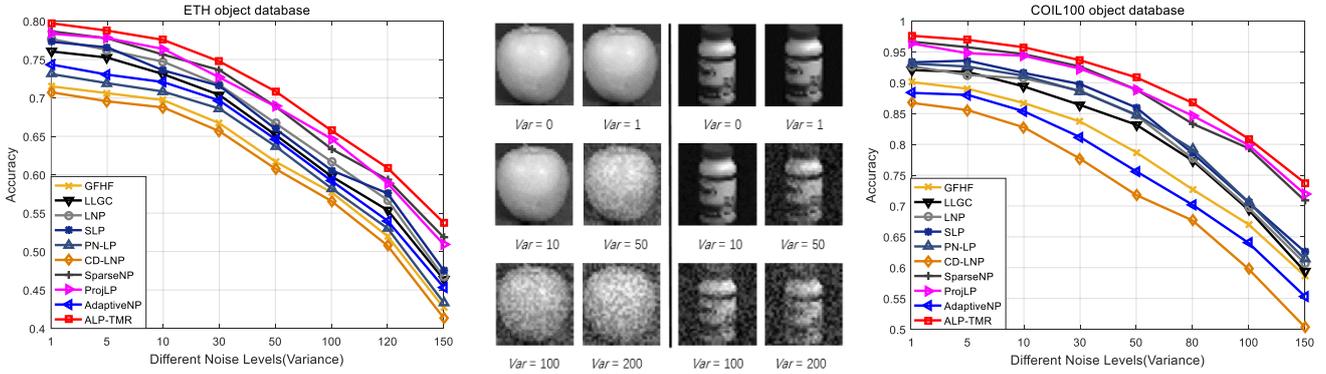

**Fig. 15:** Classification results of each algorithm under various levels of pixel corruption on ETH80 (left) and COIL100 (right).

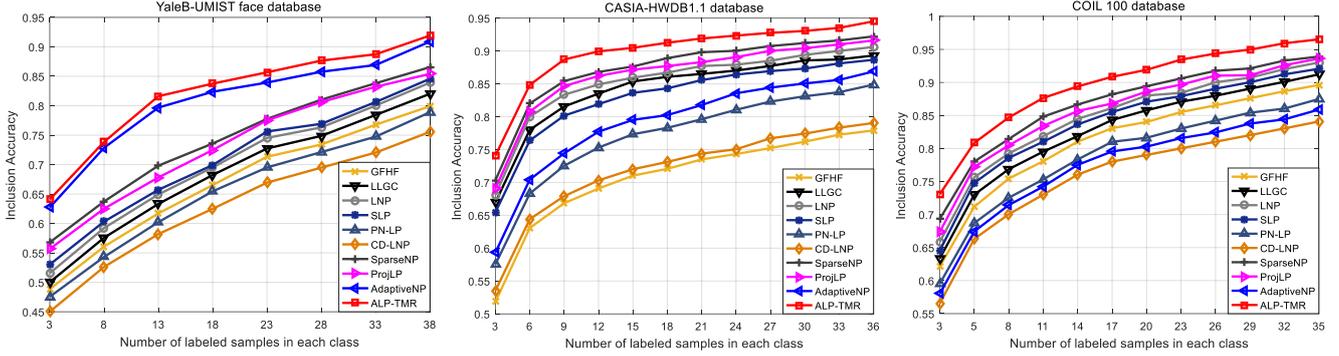

**Fig. 16:** Out-of-sample data inclusion performances on YaleB-UMIST (left), CASIA-HWDB1.1 (middle) and COIL100 (right).

The averaged classification results over 15 random splits that are based on noisy databases are presented in Figs. 13-15. We make the following observations: 1) the classification performance of each method decreases as the level of noise increases; 2) ALP-TMR outperforms its competitors by using the triple matrix recovery to enhance the robustness by removing the noise from the data, the predicted soft labels and the similarity jointly; and 3) AdaptiveNP obtains comparable results to ALP-TMR for face recognition but delivers poorer results for object and handwriting digit recognition. ProjLP and SparseNP outperform the remaining methods since they also used feasible methods to reduce the unfavorable mixed signs and noise from the estimated soft labels.

### E. Out-of-Sample Extension for Handling New Data

We discuss ALP-TMR to consider out-of-sample new data. We apply a similar inclusion approach to LNP via label reconstruction [2]. However, our approach differs from LNP as follows: Our method reconstructs the label of each new sample $x_{new}$ by using the recovered clean soft labels $F - E_F$ of the training set, whereas LNP uses the original soft label matrix $F$ of the training set for the label reconstruction.

For specified new test data $x_{new}$, our method searches its $K$-neighbors from the labeled training set and calculates the weight vector $w(x_{new}, x)$ that specifies the contributions of the neighbors to the reconstruction of $x_{new}$. Analogous to LNP [2], we apply the following smoothness criterion for $x_{new}$:

$$\vartheta(f_{new}) = \sum_{i: x_i \in X_L, x_i \in \mathbb{N}(x_{new})} w(x_{new}, x_i)(f_{new} - (F - E_F)_i)^2 \\ = \sum_{i: x_i \in X_L, x_i \in \mathbb{N}(x_{new})} w(x_{new}, x_i)(f_{new} - f_i)^2 \quad (31)$$

where $\mathbb{N}(x_{new})$ contains the $K$-nearest neighbors of $x_{new}$ and $f_i$ is the $i$-th column of the recovered clean soft label matrix $F$. Since $\vartheta(f_{new})$ is convex in $f_{new}$, it is minimized when

$$f_{new} = \sum_{i: x_i \in X_L, x_i \in \mathbb{N}(x_{new})} w(x_{new}, x_i) f_i. \quad (32)$$

Then, the label of sample $x_{new}$ can be reconstructed using the soft labels of the labeled data in the training set:

$$f_{new} = \min_{f_{new}} \left\| f_{new} - \sum_{i: x_i \in X_L, x_i \in \mathbb{N}(x_{new})} w(x_{new}, x_i) f_i \right\|^2, \quad (33)$$

from which the soft label vector $f_{new}$ of each $x_{new}$ can be estimated by minimizing Eq. (33), where the largest element in $f_{new}$ determines the class assignment of sample $x_{new}$.

We also conduct a simulation to evaluate the performance of ALP-TMR on including out-of-sample new data. Three types of real images, namely, YaleB-UMIST faces, CASIA-HWDB1.1 handwriting digits and COIL100 objects, are considered. The same label reconstruction scheme as LNP [2] is adopted for GFHF, LLGC, LNP, PN-LP, SLP, CD-LNP, ProjLP, SparseNP and AdaptiveNP. In this study, half of the labeled images from each class are selected to be corrupted and the percentage of corruptions is set to 0.5. The number of nearest neighbors is fixed to seven for each algorithm for fair comparison. We present the classification results in Fig. 16, where the numbers of training data points per subject class are set to 80, 190 and 60 for YaleB-UMIST, CASIA-HWDB1.1 and COIL100, respectively, and the remaining data points are considered as unlabeled samples. We find that the inclusion performance of each method can be improved by increasing the number of labeled training samples. ALP-TMR outperforms the other compared methods by delivering superior performance. ProjLP and SparseNP also perform well on each database since they have also considered removing noise from the predicted soft labels of the training samples. AdaptiveNP obtains comparable results to our ALP-TMR on YaleB-UMIST; however, it performs poorly on the CASIA-HWDB1.1 and COIL100 databases. CD-LNP obtains the worst results in most cases, and the remaining methods perform comparably with one another.

## F. Comparison to Deep Convolutional Networks

In this study, we compare the classification results that are based on deep convolutional features [47]. Three representative deep network models, namely, VGG16 [46], VGG19 [46] and Alexnet [47], are employed for the evaluations. Since deep network models perform classification based on deep convolutional features [59], while the classification process of ALP-TMR is conducted on original data, direct comparison between them is unfair. In this study, we compare them as follows. First, we use the VGG16, VGG19 and Alexnet models to extract the deep features from original images. In this step, the sizes of the original images are set to the required input sizes of the VGG16, VGG19 and Alexnet models. Since the inputs of the deep network frameworks are RGB images, in this study, we convert each gray image into an RGB image by copying the gray image into each of the three channels of RGB space. Features from the penultimate full connection layer ('fc7') are chosen for evaluations, where the dimension of the features is 4096. Then, we construct a feature matrix based on the extracted deep features. Second, we perform the classification task based on the constructed deep feature matrix by employing various classifiers for fair comparison. Based on the same feature matrix, we compare the classification results of ALP-TMR with those of two widely used classifiers, namely, multiclass *Support Vector Machine* (SVM, one-against-one) [49] and *Softmax* [48]. The accuracies of all the classifiers are averaged based on eight operations to avoid bias for fair comparison.

In this simulation, two popular datasets are employed: CMU PIE and USPS. For classification, we select 10 and 30 feature vectors from the deep feature matrix for each class as labeled training sets for CMU PIE, and we select 20 and 50 feature vectors as labeled training sets for USPS. For semi-supervised classification via ALP-TMR, the number of unlabeled samples is set to the same as that of labeled data. The comparison results over three deep networks are presented in TABLEs VIII, IX and X. According to the results, our ALP-TMR realizes higher accuracy than SVM and Softmax based on the same deep features by discovering the underlying structures that are hidden in both labeled and unlabeled data. Hence, our ALP-TMR can also perform well using convolutional features, which encourages us to incorporate the label propagation into deep learning in future work.

TABLE VIII:
PERFORMANCE COMPARISON OVER ALEXNET.

| Dataset / Method | CMU PIE face | | USPS digits | |
|---|---|---|---|---|
| | *10 label* | *30 label* | *20 label* | *50 label* |
| Alexnet + SVM | 60.50 | 82.06 | 87.07 | 96.87 |
| Alexnet + Softmax | 77.36 | 93.64 | 89.19 | 96.40 |
| Alexnet + ALP-TMR | **79.51** | **93.79** | **90.34** | **97.53** |

TABLE IX:
PERFORMANCE COMPARISON OVER VGG16.

| Dataset / Method | CMU PIE face | | USPS digits | |
|---|---|---|---|---|
| | *10 label* | *30 label* | *20 label* | *50 label* |
| VGG16 + SVM | 59.90 | 80.07 | 81.41 | 93.21 |
| VGG16 + Softmax | 74.85 | 89.99 | 81.66 | 92.94 |
| VGG16 + ALP-TMR | **80.41** | **94.18** | **85.82** | **95.34** |

TABLE X:
PERFORMANCE COMPARISON OVER VGG19.

| Dataset / Method | CMU PIE face | | USPS digits | |
|---|---|---|---|---|
| | *10 label* | *30 label* | *20 label* | *50 label* |
| VGG19 + SVM | 56.57 | 78.02 | 79.90 | 92.24 |
| VGG19 + Softmax | 72.51 | 88.46 | 83.30 | 92.27 |
| VGG19 + ALP-TMR | **76.79** | **92.86** | **84.38** | **93.49** |

## IV. REMARKS AND DISCUSSION

In this section, the reasons for categorizing our present work into the areas of neural networks and learning systems (NNLS) and the relation between our semi-supervised learning model and deep neural networks are described as follows:

(1) Our proposed approach is a semi-supervised learning system, which has the same objectives as a neural network learning system, namely, representation and classification. Our proposed method aims at learning a discriminative representation and predicting labels of samples via label propagation for image classification. The major difference is that almost all available neural network learning systems perform classification in a supervised learning manner, whereas our approach is an effective semi-supervised learning method. Semi-supervised learning for representation and classification is a very hot topic in the areas of NNLS, such as in the recently published TNNLS papers [61-70], in which a similar topic on semi-supervised learning to our work is studied.

(2) Our approach is an effective semi-supervised classifier. Hence, it can be potentially used as a classifier in the output layer of a deep neural network architecture to classify samples. Since our proposed ALP-TMR can use a small amount of labeled data and a large amount of unlabeled data for semi-supervised learning to predict the labels of samples, our work will have a wider range of application areas than the widely used fully supervised classifiers in the output layer, e.g., *multi-class* SVM [49] and *Softmax* [48]. SVM and Softmax require that all the training samples be labeled, while the amount of labeled data are typically limited in practice and the labeling process is costly. By discovering the intrinsic relations of labeled and unlabeled data, the classification results can be potentially improved via our semi-supervised method over SVM and Softmax based on the same deep features, according to the results in *Subsection F* of Section IV. Hence, our ALP-TMR can also perform well using convolutional features, which encourages us to incorporate the LP process into the deep learning models in future work.

In conclusion, our approach is a semi-supervised learning system, and our algorithm can be used as a classifier in the output layers of deep network models.

## V. CONCLUSIONS AND FUTURE WORK

We have explored the robust semi-supervised classification problem for predicting the labels of samples. An effective triple-matrix-recovery-driven robust auto-weighted label-propagation framework is proposed. To improve the accuracy and performance of the similarity measure, representation learning and semi-supervised classification, we have presented a simple yet effective triple matrix decomposition mechanism for recovering the underlying clean data, clean soft labels and clean weight spaces jointly from the original data, the predicted labels and the weights, which typically have unfavorable mixed signs or incorrect inter-class connections. Thus, our model potentially delivers enhanced data representation and classification results by performing robust label prediction on recovered clean data and weight spaces.

We have evaluated our method via extensive simulations. By visualizing the recovered clean soft labels and the recovered clean weights, we conclude that the proposed triple matrix recovery mechanism can indeed lead to more accurate similarity measures and data representations and more discriminating predicted labels. Quantitative classification results on public face, handwriting and object databases also demonstrate that our method can realize enhanced label prediction performance compared with related algorithms. In the future, we will incorporate the data inclusion process to form a unified model that can handle new data directly. The extension of our method to the other emerging applications, such as image retrieval and annotation, also merits investigation.

ACKNOWLEDGMENTS

This work is partially supported by National Natural Science Foundation of China (61672365, 61732008, 61725203, 61622305, 61871444, 61806035), High-Level Talent of the "Six Talent Peak" Project of Jiangsu Province of China (XYDXX-055), and the Fundamental Research Funds for the Central Universities of China (JZ2019HGPA0102). Dr. Zhao Zhang is the corresponding author of this paper.

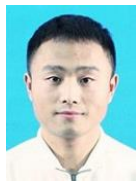

**Huan Zhang** is working toward a graduate degree at School of Computer Science and Technology, Soochow University, Suzhou, P. R. China. His current research interests mainly include data mining, machine learning and pattern recognition. He is interested in designing robust and effective semi-supervised learning methods for pattern classification.

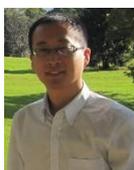

**Zhao Zhang** (SM'17- ) is a Full Professor in the School of Computer Science & School of Artificial Intelligence, Hefei University of Technology, Hefei, P. R. China. He received a Ph.D. degree from the Department of EE at City University of Hong Kong in 2013. Dr. Zhang was a Visiting Research Engineer at the Learning & Vision Research Group, National University of Singapore, from Feb. to May 2012. Then, he visited the National Laboratory of Pattern Recognition (NLPR) at Chinese Academy of Sciences (CAS), from Sept. to Dec. 2012. From Oct. 2013 to Oct. 2018, he was a Distinguished Associate Professor at the School of Computer Science and Technology, Soochow University, Suzhou, China. His current research interests include Multimedia Data Mining & Machine Learning, Image Processing & Computer Vision. He has authored/co-authored over 80 technical papers published at prestigious journals and conferences, such as IEEE TIP (4), IEEE TKDE (6), IEEE TNNLS (8), IEEE TSP, IEEE TCSVT, IEEE TCYB, IEEE TBD, IEEE TII (2), ACM TIST, Pattern Recognition (6), Neural Networks (8), Computer Vision and Image Understanding, ACM Multimedia, IJCAI, ICDM (4), ICASSP and ICMR, etc. Specifically, he has published 24 regular papers in IEEE/ACM Transactions. Dr. Zhang is serving/served as an Associate Editor (AE) for IEEE Access, Neurocomputing and IET Image Processing. Besides, he has been acting as a Senior PC member/Area Chair of ECAI、BMVC、PAKDD and ICTAI, and a PC member for 10+ popular prestigious conferences (e.g., CVPR、ICCV、IJCAI、AAAI、ACM MM、ICDM、CIKM and SDM). He is now a Senior Member of the IEEE, and a Senior Member of the CCF. He is a Senior Member of the IEEE.

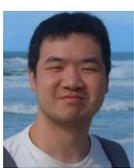

**Mingbo Zhao** (M'13-) received a Ph.D. degree in computer engineering from Department of Electronic Engineering at City University of Hong Kong in 2013. Dr. Zhao is currently a Senior Research Fellow at the Department of Electronic Engineering, City University of Hong Kong, Hong Kong SAR. His current research interests include machine learning, data mining, pattern recognition and applications.

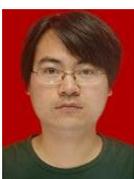

Qiaolin Ye (M'15) received a Ph.D. degree in pattern recognition and intelligence system from the Nanjing University of Science and Technology, Nanjing, in 2013. He is currently an Associate Professor of the Department of Computer Science with Nanjing Forestry University. He has authored over 50 papers in pattern recognition and machine learning. Some are published in IEEE TNNLS, IEEE TIFS, IEEE TIP, and IEEE TSCVT. His research interests include machine learning and data mining.

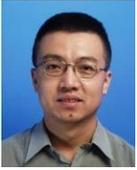

**Min Zhang** is a Distinguished Professor at the Soochow University (China). He received his bachelor's degree and Ph.D. degree from the Harbin Institute of Technology in 1991 and 1997, respectively. His current research interests include machine learning, intelligent computing and natural language processing. He has co-authored over 130 papers in leading journals and conferences, and has co-edited 10 books that were published by Springer and IEEE. He has been actively contributing to the research community by organizing many conferences as a conference chair, program chair and organizing chair and by giving talks at many conferences and lectures.

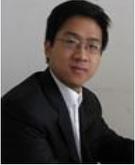

**Meng Wang** is a professor in the Hefei University of Technology, China. He received a B.E. degree and a Ph.D. degree in the Special Class for the Gifted Young from the University of Science and Technology of China, Hefei, China. He worked as an associate researcher at Microsoft Research Asia and as a core member in a startup in the Bay area. After that, he worked as a senior research fellow in National University of Singapore. His current research interests include multimedia content analysis, search, mining, recommendation, and large-scale computing. He has authored over 100 journal and conference papers in these areas, including in TMM, TNNLS, TCSVT, TIP, TOMCCAP, ACM MM, WWW, SIGIR, and ICDM. He received paper awards from ACM MM 2009 (Best Paper Award), ACM MM 2010 (Best Paper Award), MMM 2010 (Best Paper Award), ICIMCS 2012 (Best Paper Award), ACM MM 2012 (Best Demo Award), ICDM 2014 (Best Student Paper Award), PCM 2015 (Best Paper Award), SIGIR 2015 (Best Paper Honorable Mention), IEEE TMM 2015 (Best Paper Honorable Mention), and IEEE TMM 2016 (Best Paper Honorable Mention). He is the recipient of the ACM SIGMM Rising Star Award 2014. He is/has been an Associate Editor of IEEE Trans. on Knowledge and Data Engineering (TKDE), IEEE Trans. on Neural Networks and Learning Systems (TNNLS) and IEEE Trans. on Circuits and Systems for Video Technology (TCSVT).